\documentclass{article}

\usepackage{arxiv}

\usepackage[utf8]{inputenc} 
\usepackage[T1]{fontenc}    
\usepackage{hyperref}       
\usepackage{url}            
\usepackage{nicefrac}       
\usepackage{microtype}      
\usepackage{lipsum}

\usepackage{booktabs} 
\usepackage{xcolor} 
\usepackage{colortbl} 
\usepackage{graphicx} 
\usepackage{array} 
\usepackage{multirow} 
\usepackage{makecell} 
\usepackage{amsmath,amsfonts} 
\usepackage{pifont} 

\title{Proto-FG3D: Prototype-based Interpretable Fine-Grained 3D Shape Classification}

\author{
Shuxian Ma \\
  School of Information Science and Engineering\\
  University of Jinan\\
  Jinan  250022, China \\
  \texttt{mashuxian@stu.ujn.edu.cn} \\
   \And
Zihao Dong \\
  School of Information Science and Engineering\\
  University of Jinan\\
  Jinan  250022, China \\
  \texttt{ise\_dongzh@ujn.edu.cn} \\
  \And 
Runmin Cong \\
  School of Control Science and Engineering\\
  Shandong University\\
  Jinan 250061, China \\
  \texttt{rmcong@sdu.edu.cn} \\
  \And
Sam Kwong \\
  Department of Computing and Decision Sciences\\
  Lingnan University\\
  Hong Kong SAR, China \\
  \texttt{samkwong@ln.edu.hk} \\
  \And
Xiuli Shao \\
  College of Computer Science\\
  Nankai University\\
  Tianjin  300350, China \\
  \texttt{shaoxl@nankai.edu.cn} \\
}

\begin{document}
\maketitle
\begin{abstract}
Deep learning-based multi-view coarse-grained 3D shape classification has achieved remarkable success over the past decade, leveraging the powerful feature learning capabilities of CNN-based and ViT-based backbones. However, as a challenging research area critical for detailed shape understanding, fine-grained 3D classification remains understudied due to the limited discriminative information captured during multi-view feature aggregation, particularly for subtle inter-class variations, class imbalance, and inherent interpretability limitations of parametric model. To address these problems, we propose the first prototype-based framework named Proto-FG3D for fine-grained 3D shape classification, achieving a paradigm shift from parametric softmax to non-parametric prototype learning. Firstly, Proto-FG3D establishes joint multi-view and multi-category representation learning via \textit{Prototype Association}. Secondly, prototypes are refined via \textit{Online Clustering}, improving both the robustness of multi-view feature allocation and inter-subclass balance. Finally, prototype-guided supervised learning is established to enhance fine-grained discrimination via prototype-view correlation analysis and enables ad-hoc interpretability through transparent case-based reasoning. Experiments on FG3D and ModelNet40 show Proto-FG3D surpasses state-of-the-art methods in accuracy, transparent predictions, and ad-hoc interpretability with visualizations, challenging conventional fine-grained 3D recognition approaches.
\end{abstract}


\section{Introduction}
\label{sec:intro}

Learning a shape representation from multiple rendered views is essential for understanding 3D content. Compared to point-based and voxel-based methods, view-based 3D shape classification models achieve superior accuracy and efficiency by projecting 3D shapes into multi-view 2D images. CNN-based approaches have significantly enhanced feature aggregation through hierarchical view weighting \cite{2015mvcnn, 2018gvcnn, 2018MVCNNnew, 20193d2seqviews, 2020SMVCNN} or joint optimization of viewpoint estimation \cite{2018veram, 2018seqviews2seqlabels, 2019mlvcnn, 20193d2seqviews, 2020viewGCN}. Due to the superior capability of Transformers in discriminating subtle geometric patterns, ViT-based frameworks \cite{2021CVR, 2021mvt, 2024vsformer, 2024gmvit} have been achieved by leveraging self-attention mechanisms to dynamically fuse multi-view features. To ensure interpretability in the aforementioned representation learning paradigm for multi-view image-based 3D shape classification, activation maximization \cite{Simonyan2013visual, nguyen2016visual,carter2019activation} helps identify view-specific discriminative patterns. The rise of ViT-based architectures enables self-attention mechanisms to reflect feature prioritization across views. Attention maps \cite{2022hmtn, 2024gmvit, 2024mhsan, 2024vsformer} further expose cross-view interactions. However, these view-based methods with limited interpretability still exhibit deficiencies in learning the small variance among subcategories.

Driven by growing real-world requirements, fine-grained 3D shape recognition has emerged as a critical task for differentiating subtle subcategory variations within object categories. The FG3D dataset \cite{liu2021fine} introduces a 3D shape benchmark capturing fine-grained structural details via multi-view rendering. While significant advances have been achieved in coarse-grained 3D recognition, fine-grained classification continues to face persistent challenges including: \textbf{(i) Local Fine-Grained Inter-Class Discrepancies:} Subtle distinctions between subcategories primarily arise from part-level variations in identical object classes, highlighting the necessity of precise multi-view differentiation to optimize model performance. \textbf{(ii) Severe Subclass-Imbalance:} The FG3D dataset exhibits significant class imbalance, with the training sample distribution ranging from 2,027 instances for sedans to merely 14 for tricycles. \textbf{(iii) Weak Interpretability of Fine-Grained Visual Relationships:} Fine-grained 3D classification places higher demands on model interpretability, requiring transparent and reliable reasoning processes rather than black-box decision making.

\begin{figure}[tbhp]
\begin{center}
\includegraphics[width=0.85\textwidth]{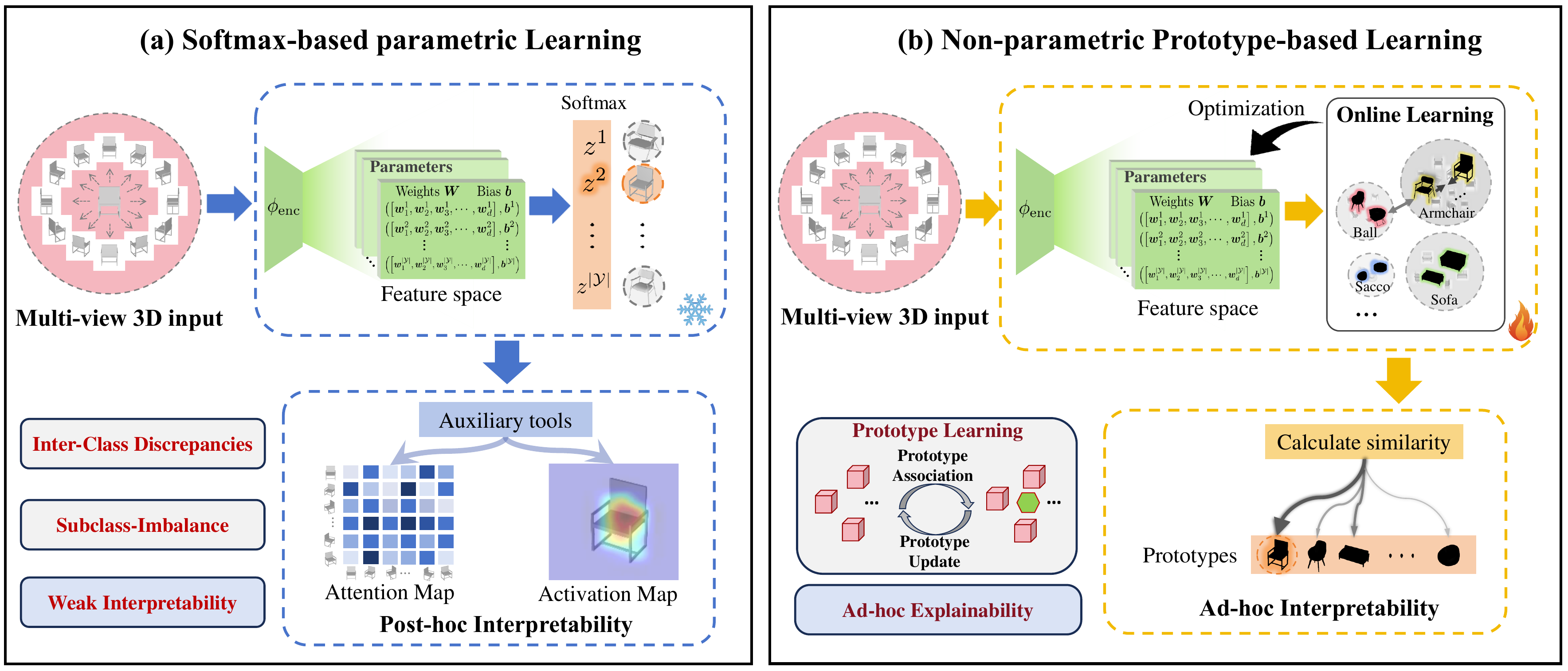}
\caption{Multi-view 3D shape classification paradigms: (a) Parametric softmax can be interpreted as a learnable prototype-based approach, where class- and view-coupled prototypes are learned in a fully parametric manner. (b) Non-parametric prototype learning directly identifies subcluster centers of embedded features as prototypes, enabling per-view predictions through nonparametric nearest prototype retrieval.} 
\label{fig:Intro_Proto}
\end{center}
\end{figure}

To systematically tackle the challenges of 3D fine-grained classification, a critical question becomes more fundamental: \textbf{What unified representation processing is applied to the features extracted by backbones to achieve efficient, reliable and human-controllable classification processes?} As illustrated in Fig. \ref{fig:Intro_Proto} (a), CNN-based and ViT-based methods typically rely on learned high-dimensional decision boundaries that map features to class probabilities via fully connected layers. Effective for coarse-grained tasks, their opacity necessitates post-hoc interpretability tools, causing a disconnect between model reasoning and explanations. Minsky introduced "Prototype-Derived Patterns", asserting that character recognition relies on comparing inputs to fixed prototypes \cite{minsky1961steps}, a theory supported by cognitive psychology \cite{newell1972human} and neuroscience \cite{kriegeskorte2008matching}, which show the human visual system recognizes objects by matching them to memorized prototypes in the perirhinal cortex. Inspired by these insights, we propose a non-parametric prototype-based learning framework to rethink 3D fine-grained classification, as shown in Fig. \ref{fig:Intro_Proto} (b). The proposed Proto-FG3D projects 3D objects to multi-view 2D projections, encodes them via a shared backbone, and clusters features into learnable prototypes capturing discriminative subcategory patterns. Prototypes are optimized through online clustering and momentum updating to ensure balanced groupings. During training, the framework integrates hybrid feature-space optimization, bridging metric learning and representation learning. At inference, predictions are made by comparing input features to their nearest prototypes.

The proposed framework focuses on three key advantages: \textbf{(i) Enhanced Fine-Grained Discrimination:} Unlike conventional parametric classifiers aggregating multi-view features into opaque embeddings, our method explicitly models discriminative cross-view patterns though learnable exemplars. Prototypes integrate features from diverse viewpoints to capture representations, preserving structural continuity and ensuring view invariance and consistency. \textbf{(ii) Robustness to Class Imbalance:} Prototypes for rare subcategories are initialized with cluster centroids and conservatively updated to prevent dominance by frequent classes. This regularization reduces the gap between Average Instance Accuracy (AIA) and Average Class Accuracy (ACA). \textbf{(iii) Ad-Hoc Interpretability Aligned with Human Reasoning: } The framework generates self-explanable predictions by leveraging class-specific prototypes as human-understandable exemplars through case-based reasoning.

\section{Related work}
\noindent \textbf{3D Classification: From CG to FG.$\,$} Existing view-based \textbf{coarse-grained (CG)} 3D classification methods, including CNN-based \cite{2015mvcnn, 2018gvcnn}, LSTM-based \cite{2018veram}, RNN-based \cite{2018seqviews2seqlabels} and ViT-based architectures \cite{2021mvt, 2024vsformer}, leverage projected 2D images to achieve state-of-the-art performance in large-scale 3D shape recognition tasks. These fully parametric DNNs fail to preserve fine geometric details due to their reliance on statistical approximations of 3D structural relationships. \textbf{Fine-grained (FG)} 3D classification demands precise discrimination of subtle geometric variations, which challenged by viewpoint heterogeneity and structural complexity. The FG3D-Net method \cite{liu2021fine} introduces large-scale fine-grained recognition and integrates semantic parts effectively, its significant accuracy gaps between ACA and AIA metrics highlight the need for a novel paradigm to capture subtle inter-class geometric variations while mitigating class imbalance.

\noindent \textbf{Prototype Learning.$\,$} Prototype-based learning provides an interpretable framework for classification by aligning input embeddings with class-specific prototypes. In supervised classification \cite{wu2018improving, yang2018robust, wang2022visual}, modern implementations first map input samples to deep features and then classify them via prototype similarity, demonstrating robustness in both few-shot \cite{snell2017prototypical, mettes2019hyperspherical,tian2020rethinking} and zero-shot learning \cite{jetley2015prototypical, xu2020attribute}.
Beyond basic classification tasks, this paradigm has been effectively extended to semantic segmentation \cite{wang2019panet, li2021adaptive, feng2023clustering, zhou2024prototype} through prototype-based pixel-wise matching. Unsupervised learning frameworks \cite{wu2018unsupervised, li2020prototypical, ye2020augmentation} have also adopted prototype-guided clustering to uncover latent patterns without labeled data. The inherent interpretability of prototype matching further drives the development of interpretable networks \cite{li2018deep, feng2024interpretable3d}, addressing the growing demand for transparent decision-making.
However, most prior work relies on class-specific local representations, neglecting coupled global analysis of multi-feature interactions. 

\noindent \textbf{Post-hoc \& Ad-hoc Interpretability.$\,$}
Most previous research on 3D interpretability focuses on \textbf{post-hoc explanations} from parametric softmax-based DNNs. One approach uses activation maximization \cite{Simonyan2013visual,nguyen2016visual,carter2019activation} to identify input patterns. Another method \cite{zhou2016learning,selvaraju2016grad,chefer2021transformer,zech2018variable} localizes discriminative features through saliency-driven ROI extraction or learned attention mapping. Recent multi-view approaches like VSFormer \cite{2024vsformer}, MHSAN \cite{2024mhsan}, and DAN \cite{2021dan} leverage attention maps, while MV-HFMD \cite{2024mvhfmd} employs activation maps. However, post-hoc explanations remain unreliable, introducing unknown discrepancies between explanations and actual model decisions. \textbf{Ad-hoc interpretability} integrates more interpretable mechanisms into the design of black-box DNNs.
Recent developments \cite{chen2019looks, xue2024protopformer} in ad-hoc interpretability have been made in the context of 2D image classification, demonstrating the effectiveness of prototypical and non-parametric approaches to enhance recognition's explainability.
Building on these advancements, \cite{feng2023clustering} applied clustering-based algorithm for 3D point clouds. However, ad-hoc interpretability for 3D multi-view fine-grained classification remains underexplored in DNN-based decision systems. 

\section{Prototype-based learning for FG3D} 

\begin{figure}[tbhp]
\begin{center}
\includegraphics[width=0.88\textwidth]{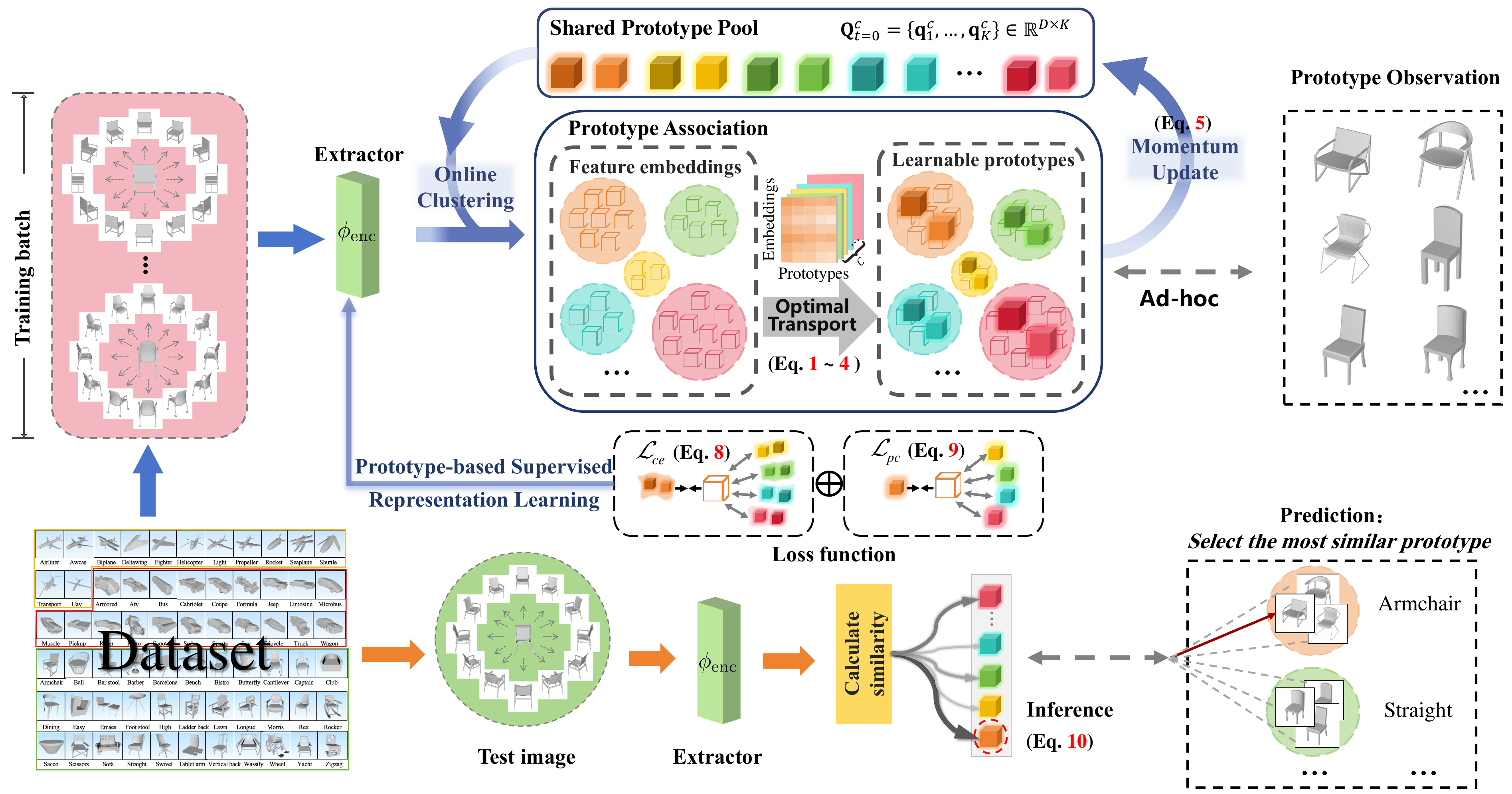}
\caption{Architecture of prototype-based fine-grained 3D shape classification model. The visualization of initial prototypes (e.g., \( \mathbf{Q}_{t=0}^c \)) is reserved for future experimental results. }
\label{fig:Proto-Arch}
\end{center}
\end{figure}

\noindent \textbf{Overall Architecture.$\,$} As illustrated in Fig. \ref{fig:Proto-Arch}, the proposed online clustering framework comprises a shared encoder $f$, a prototype pool $\mathbf{P}$, and a prototype association mechanism. The shared encoder extracts feature embeddings $\mathbf{H} = \{ \mathbf{h}_j \in \mathbb{R}^D \}_{j=1}^{V}$ from $V$ input views, where $D$ denotes the feature dimension. To enable discriminative feature learning, we maintain a shared pool of $K$ trainable prototypes per class, denoted as $\mathbf{Q}^c = \{ q_k^c \}_{k=1}^K \in \mathbb{R}^{D \times K}$, and formulate an optimal transport problem to softly assign each feature embedding $z_j$ to these prototypes as a probability distribution $\mathbf{q}_j \in \mathbb{R}^K$ $(\|\mathbf{q}_j\|_2 = 1)$. These class-specific prototypes establish multi-view correlations through \textbf{Prototype Association} during training, with the shared prototype pool being progressively refined via exponential moving averages of assigned features (i.e., \textbf{Prototype Update}). The model is optimized using a joint objective function that enforces feature-prototype similarity maximization while separating dissimilar pairs (i.e., \textbf{Prototype-based Supervised Representation Learning}). During inference, the final prediction for a test sample is determined by retrieving the nearest prototype in the shared pool based on the learned feature embeddings.

\subsection{Prototype Association}
\noindent \textbf{Prototype Initialization.$\,$} To address the limitations of  softmax-based methods, we propose a prototype-driven learning framework that enhances both discriminability and interpretability via deterministic clustering in the representation space $\mathcal{H}$. For each class $c$, we initially cluster $V^c$ multi-view representations $\{\mathbf{h}_v^c\}_{v=1}^{V^c} \in \mathbb{R}^{D}$ into $K$  class-specific sub-centroids $\{\mathbf{q}_k^c \}_{k=1}^{K} \in \mathbb{R}^{D}$. These prototypes are then stored in a dynamic prototype pool, where they encapsulate representative multi-view patterns to capture fine-grained properties of class $c$.

\noindent \textbf{Prototype Assignment.$\,$} Traditional prototypical approaches eliminate the need for gradient-based prototype assignment by relying on manually defined mappings between prototypes and class labels \cite{rymarczyk2021protopshare} or hierarchical tree nodes \cite{nauta2021neural}. In contrast, Proto-FG3D introduces a dynamic, learnable prototype distribution mechanism enabling adaptive soft assignment from a shared prototype pool. For class $c$, we cast the assignment of  feature descriptors  $\mathbf{H}^c$ extracted from $V^c$ views to prototypes as an \textbf{optimal transport problem}. Specifically, the goal of this task is to map the multi-view images $\{I_v\}_{v=1}^{V^c}$ to $K$ class-specific prototypes. We need to find a view-to-prototype assignment matrix $\mathbf{Z}^c \in [0,1]^{K \times {V^c}}$ that maximizes the aggregated similarity between feature embeddings and class prototypes, while widening the gap between maximal and average activation to prevent prototype overcrowding in non-discriminative regions. This procedure  can be formulated as the relaxed Binary Integer Program (BIP) using soft assignments with an entropy regularization:

\begin{equation}
\small
    \mathbf{Z}^{c*} = \mathop{\text{max}}_{\mathbf{Z}^c} \left\{ \underbrace{\text{Tr}\left((\mathbf{Z}^c)^\top (\mathbf{Q}^c)^\top \mathbf{H}^c\right)}_{\text{Similarity Maximization}} + \underbrace{\kappa \cdot H(\mathbf{Z}^c)}_{\text{Regularization}} \right\} - \mathop{\text{mean}}_{\mathbf{Z}^c} \left\{ \underbrace{\text{Tr}\left((\mathbf{Z}^c)^\top (\mathbf{Q}^c)^\top \mathbf{H}^c\right)}_{\text{Similarity Maximization}} + \underbrace{\kappa \cdot H(\mathbf{Z}^c)}_{\text{Regularization}} \right\},
\label{equ:Proto_assignV2}
\end{equation}

\noindent where $\text{Tr}(\cdot)$ denotes the matrix trace. To guarantee that $\mathbf{Z}^c$ effectively represents all classes, we enforce two key constraints, including the \textbf{uniqueness constraint} and \textbf{equipartition constraint}, to ensure that each feature is assigned to exactly one prototype while maintaining balanced assignments across all prototypes. thereby preventing trivial solutions where all point samples collapse into a single prototype:

\begin{equation}
    (\mathbf{Z}^c)^\top \mathbf{1}_K = \mathbf{1}_{V^c}, \; \mathbf{Z}^c \mathbf{1}_{V^{c}} = \frac{V^{c}}{K} \mathbf{1}_K,
\label{equ:ConstV1}
\end{equation}

\noindent where $\mathbf{1}_K$ and $\mathbf{1}_{V^c}$ denote all-ones tensors of dimensions $K$ and $V^c$ respectively. In Eq. \ref{equ:Proto_assignV2}, $\kappa > 0$ is a hyperparameter that balances distribution smoothness against fidelity to the original transport problem. The entropy regularization term enhances optimization efficiency through two mechanisms: (i) \textbf{ensuring a unique solution} via strong convexity and (ii)  \textbf{stabilizing prototype computations} by preventing overly sharp predictions or degenerate assignments:

\begin{equation}
    H(\mathbf{Z}^c) = -\sum_{k=1}^{K}\sum_{v=1}^{V^c} z_{k,v}^c \log z_{k,v}^c.
\label{equ:Entropy_reg}
\end{equation}

The resulting optimization is then efficiently solved via the fast Sinkhorn-Knopp algorithm \cite{distances2013lightspeed} and APDAGD \cite{lin2019efficient}, expressed as a normalized exponential matrix:

\begin{equation}
    \mathbf{Z}^{c*} = \text{diag}(\mathbf{\mu}) \exp\left(\frac{(\mathbf{Q}^c)^\top \mathbf{H}^c}{\kappa}\right) \text{diag}(\mathbf{\nu}),
\label{equ:SK_op}
\end{equation}

\noindent where $\mathbf{\mu}$ and $\mathbf{\nu}$ are learnable normal feature vectors, which can be efficiently computed by the Sinkhorn-Knopp iteration with minimal matrix-vector multiplications. 

\subsection{Online Prototype Update}

To avoid the computational expense of offline sub-centroid estimation (requiring full-dataset class-wise clustering despite the efficiency of Sinkhorn-Knopp iteration in Eq. \ref{equ:SK_op}), we employ exponential moving average (EMA)-based updating with online clustering to dynamically capture the local structure in multi-view feature space. Each training step conducts class-specific clustering on the current view set, followed by incremental sub-centroid updates:

\begin{equation}
    \mathbf{q}_k^c \leftarrow \eta^t \mathbf{q}_k^c(0) + (1 - \eta)\sum_{i=1}^{t} \eta^{t-i} \bar{\mathbf{H}}_k^c(i),
\label{equ:EMA_update}
\end{equation}

\noindent where the momentum coefficient $\eta$ implements an exponentially decaying weight allocation over time $t$. Specifically, $\eta^0=0.999$ ensures rapid convergence in the initial stage, while for $t>T$ it dynamically adjusts as $\eta_t=\min(0.999, 1-\frac{1}{t+1})$ to achieve progressive decay. $\bar{{\mathbf{H}}}_k^c \in \mathbb{R}^D$ is the mean vector of $\{\mathbf{h}_{i}^c\}_{i=1}^{V^c}$, defined as:

\begin{equation}
   \bar{\mathbf{H}}_k^c = \frac{1}{|\mathcal{S}_k^c|} \sum_{\mathbf{h^c} \in \mathcal{S}_k^c} \frac{1}{V^c} \sum_{i=1}^{V^c}\mathbf{h}_i^c,
\label{equ:AVG_feature}
\end{equation}

\noindent where $\mathcal{S}_k^c$ is the set of view features $\mathbf{h}^c$ assigned to the $k$-th prototype for class $c$.  In the final training epochs, to further refine the prototypes and mitigate class imbalance, the averaged embedding may be replaced by the feature vector of the nearest training sample.

\subsection{Prototype-based Supervised Representation Learning} 
\noindent \textbf{Intra-class Alignment.$\,$} To reduce intra-cluster variation, our prototype-based learning method achieves feature compact structure by directly minimizing the distance between each $\ell_2$ normalized embedded view feature and its assigned prototype:

\begin{equation}
    d(\mathbf{h}_v, \mathbf{Q}^c) = \min_{1 \leq k \leq K} \left( 1 - \mathbf{h}_v^\top \mathbf{q}_k^c \right).
\end{equation}

Therefore, the multi-view classification loss employs cross-entropy to simultaneously optimize: (i) feature-to-prototype alignment for correct classes $c$, and (ii) margin constraints against incorrect prototypes through distance metric learning:

\begin{equation}
    \mathcal{L}_{ce} = -\log \frac{\exp\left(-d(\mathbf{h}_v, \mathbf{Q}^{c})\right)}{\sum_{c' \in \mathcal{C}} \exp\left(-d(\mathbf{h}_v, \mathbf{Q}^{c'})\right)},
\end{equation}

\noindent \textbf{Inter-prototype Discrepancy.$\,$} To align view embeddings with their corresponding prototypes while maximizing the separation from irrelevant ones, we employ a view-prototype contrastive learning strategy that simultaneously: (i) attracts each embedding $\mathbf{h}_v$ to its assigned positive prototype, and (ii) repels it from negative prototypes $\mathcal{Q}^-$:

\begin{equation}
    \mathcal{L}_{pc} = -\log \frac{\exp\left(\mathbf{h}_v^\top \mathbf{q}_k^{c}/\tau\right)}{\exp\left(\mathbf{h}_v^\top \mathbf{q}_k^{c}/\tau\right) + \sum_{\mathbf{q}-\in \mathcal{Q-}} \exp\left(\mathbf{h}_v^\top \mathbf{q}-/\tau\right)},
\end{equation}

\noindent where the temperature parameter $\tau$ controls the sharpness of similarity distribution. The combined loss function over all the training multi-view samples is defined as: $\mathcal{L}_{total} = \mathcal{L}_{ce} + \alpha \mathcal{L}_{pc}$, where $\alpha$ is a hyperparameter that serves as the weighting factor to balance the contribution of intra-class compactness and inter-prototype separation.

\noindent \textbf{Prototype-based Inference.$\,$} The fine-grained prediction is obtained by nearest-prototype matching between the embedded feature $\mathbf{h}_v$ of a test sample and all learned class prototypes:

\begin{equation}
    \hat{c}_v = \mathop{\text{argmin}}_{c} \left( \min_{1 \leq k \leq K} \left( 1 - \mathbf{h}_v^\top \mathbf{q}_k \right) \right).
\end{equation}

This nearest-prototype classifier achieves robust fine-grained classification via discriminative feature-prototype alignment and ensures interpretability via transparent decisions. 

\section{Experimental Results and Analysis}

\subsection{Experimental Settings}

\noindent \textbf{Datasets.$\,$} We perform extensive experiments on two widely used 3D object classification benchmarks. \textbf{ModelNet40} \cite{wu20153d}  contains a total of 12,311 objects distributed in 40 categories, with 9,843 objects designated for training (80\%) and 2,468 for testing (20\%). In our experiments, we employ a circle-12 camera configuration, where 12 views are captured from equidistant viewpoints at a $30^\circ$ elevation relative to the centroid of the object.
\textbf{FG3D} \cite{liu2021fine} is a fine-grained 3D shape dataset that comprises three main categories: airplane, car, and chair, containing 3,441, 8,235, and 13,054 shapes divided into 13, 20, and 33 subcategories, marked by pronounced imbalance between subcategories that poses significant challenges.

\noindent \textbf{Baselines.$\,$} We mainly compare with both CNN-based (MVCNN-new \cite{2018MVCNNnew}, GVCNN \cite{2018gvcnn}, SMVCNN \cite{2020SMVCNN}) and ViT-based (DAN \cite{2021dan}, VSFormer \cite{2024vsformer}) approaches, with VSFormer currently achieving state-of-the-art performance on the ModelNet40 dataset. 

\noindent \textbf{Evaluation Protocols.} The performance of multi-view 3D classification model is assessed using two key metrics: \textbf{Average Instance Accuracy (AIA)}, measuring the percentage of correctly classified test instances, and \textbf{Average Class Accuracy (ACA)}, which computes the mean accuracy across all categories. 

\noindent \textbf{Implementation Details.} For the FG3D dataset, each 3D shape is rendered into 12 views, with the viewpoints uniformly distributed around the shape; each view is preprocessed to a size of 224 $\times$ 224 $\times$ 3. During training, views undergo random horizontal flipping with pixel normalization (mean [0.485, 0.456, 0.406], std [0.229, 0.224, 0.225]), while test views only apply normalization without any augmentation. For the ModelNet40 dataset, we follow the same 12-view rendering setup as described in prior work. 

All experiments are conducted on the same system with a 16-vCPU and a 32GB vGPU. Proto-FG3D is trained for 100 epochs with batch size 32, momentum 0.9, weight decay 0.001, and a warm-up phase of 5 epochs. The initial learning rate is 0.005 and gradually decays to 0. For the prototype-based classifier, we set the number of prototypes $K$=20, temperature parameter $\tau$=0.1, loss weight $\alpha$=0.2, and the initial momentum coefficient $\eta^0$=0.999. 

\subsection{Comparison to State-of-the-Arts} \label{sec:com_sota}

Table \ref{tb:performance FG3D} presents the fine-grained 3D shape classification performance on three categories (i.e., \textit{Airplane}, \textit{Car}, and \textit{Chair}) in the FG3D dataset and the ModelNet40 benchmark. Experiments demonstrate that multi-view models excelling in coarse-grained classification exhibit significant limitations in fine-grained tasks. Integrating our Proto-FG3D framework with iterative optimal transport algorithms (Sinkhorn [S] and APDAGD [A]) consistently improves all baselines, notably reducing the performance gap between AIA and ACA through effective class imbalance mitigation. For instance, MVCNNnew+Ours (S) achieves ACA surpassing AIA, validating superior fine-grained feature discrimination and minority-class recognition. On ModelNet40, Proto-FG3D further boosts performance across CNN and ViT-based backbones, confirming its generality in global categorical feature perception. These results further validate the prototype-based framework's robustness across granularity levels and both global/local feature extraction paradigms. 

To enhance model efficiency, a nonparametric prototype learning is proposed for multi-view discriminative representation. Our method reduces parameters and FLOPs by 1\% in both feature fine-tuning module and classification layers compared to CNN/ViT baselines. On the Airplane dataset, MVCNNnew+Ours achieves the fastest 100-epoch training in 83 minutes, whereas DAN+Ours reduces original training time by 54 minutes (26.5\%$\downarrow$), demonstrating consistent efficiency gains across various architectures. 

\begin{table*}[!htbp]
    \centering
    \tiny
    \renewcommand{\arraystretch}{1.1} 
    \caption{Performance comparison of Proto-FG3D with state of the art methods on the FG3D and ModelNet40 dataset, evaluating $K$=20 prototypes and different backbone architectures.} 
    \resizebox{\textwidth}{!}{ 
    \begin{tabular}{l|lllllllll}
    \toprule
    \multirow{2}{*}{\textbf{Method}} & \multicolumn{2}{c}{\textbf{Airplane}} & \multicolumn{2}{c}{\textbf{Car}} & \multicolumn{2}{c}{\textbf{Chair}} & \multicolumn{2}{c}{\textbf{ModelNet40}} & \multirow{2}{*}{\textbf{Time}} \\
    ~ & \textbf{AIA (\%)} & \textbf{ACA (\%)} & \textbf{AIA (\%)} & \textbf{ACA (\%)} & \textbf{AIA (\%)} & \textbf{ACA (\%)} & \textbf{AIA (\%)} & \textbf{ACA (\%)} \\
    \midrule
    FG3D-Net \cite{liu2021fine} & 93.99 & 89.44 & \textbf{79.47} & 74.03 & 83.94 & 80.04 & - & - & - \\
    \arrayrulecolor{gray!50}\midrule 
    \arrayrulecolor{black} 
    MVCNNnew \cite{2018MVCNNnew} & 93.85 & 89.16 & 76.20 & 73.25 & 83.94 & 79.58 & 93.48 & 91.73 & 1h 53min \\
    \textbf{MVCNNnew+Ours (A)} & 94.40~{\tiny\textcolor{gray}{ $\uparrow$ 0.55}} & 89.80~{\tiny\textcolor{gray}{ $\uparrow$ 0.64}} & 76.96~{\tiny\textcolor{gray}{ $\uparrow$ 0.76}} & 75.25~{\tiny\textcolor{gray}{ $\uparrow$ 2}} & 84.46~{\tiny\textcolor{gray}{ $\uparrow$ 0.52}} & 80.47~{\tiny\textcolor{gray}{ $\uparrow$ 0.89}} & 94.57~{\tiny\textcolor{gray}{ $\uparrow$ 1.09}} & 91.58~{\tiny\textcolor{gray}{ $\downarrow$ 0.15}} & -\\
    \textbf{MVCNNnew+Ours (S)} & 94.81~{\tiny\textcolor{gray}{ $\uparrow$ 0.96}} & 91.54~{\tiny\textcolor{gray}{ $\uparrow$ 2.38}} & 77.11~{\tiny\textcolor{gray}{ $\uparrow$ 0.91}} & \textbf{77.25}~{\tiny\textcolor{gray}{ $\uparrow$ 4}} & 84.56~{\tiny\textcolor{gray}{ $\uparrow$ 0.62}} & 79.33~{\tiny\textcolor{gray}{ $\downarrow$ 0.25}} & 94.57~{\tiny\textcolor{gray}{ $\uparrow$ 1.09}} & 91.58~{\tiny\textcolor{gray}{ $\downarrow$ 0.15}} & \textbf{1h 23min} \\
    \arrayrulecolor{gray!50}\midrule 
    \arrayrulecolor{black}
    GVCNN \cite{2018gvcnn} & 94.13 & 91.95 & 76.50 & 73.95 & 83.32 & 78.77 & 94.57 & 91.95 & 2h 11min \\
    \textbf{GVCNN+Ours (A)} & \textbf{95.63}~{\tiny\textcolor{gray}{ $\uparrow$ 1.5}} & \textbf{93.95}~{\tiny\textcolor{gray}{ $\uparrow$ 2}} & 78.02~{\tiny\textcolor{gray}{ $\uparrow$ 1.52}} & 75.17~{\tiny\textcolor{gray}{ $\uparrow$ 1.22}} & 84.15~{\tiny\textcolor{gray}{ $\uparrow$ 0.83}} & 80.36~{\tiny\textcolor{gray}{ $\uparrow$ 1.59}} & 95.26~{\tiny\textcolor{gray}{ $\uparrow$ 0.69}} & 92.61~{\tiny\textcolor{gray}{ $\uparrow$ 0.66}} & -\\
    \textbf{GVCNN+Ours (S)} & 95.08~{\tiny\textcolor{gray}{ $\uparrow$ 0.95}} & 93.49~{\tiny\textcolor{gray}{ $\uparrow$ 1.54}} & 77.41~{\tiny\textcolor{gray}{ $\uparrow$ 0.91}} & 76.88~{\tiny\textcolor{gray}{ $\uparrow$ 2.93}} & 84.05~{\tiny\textcolor{gray}{ $\uparrow$ 0.73}} & 80.12~{\tiny\textcolor{gray}{ $\uparrow$ 1.35}} & 95.18~{\tiny\textcolor{gray}{ $\uparrow$ 0.61}} & 92.50~{\tiny\textcolor{gray}{ $\uparrow$ 0.55}} & \textbf{1h 40min}\\
    \arrayrulecolor{gray!50}\midrule 
    \arrayrulecolor{black}
    SMVCNN \cite{2020SMVCNN} & 93.85 & 91.36 & 76.65 & 74.83 & 83.01 & 79.82 & 94.73 & 92.51 & 1h 57min\\
    \textbf{SMVCNN+Ours (A)} & 94.81~{\tiny\textcolor{gray}{ $\uparrow$ 0.96}} & 90.98~{\tiny\textcolor{gray}{ $\downarrow$ 0.38}} & 77.49~{\tiny\textcolor{gray}{ $\uparrow$ 0.84}} & 76.42~{\tiny\textcolor{gray}{ $\uparrow$ 1.59}} & 83.73~{\tiny\textcolor{gray}{ $\uparrow$ 0.72}} & 80.09~{\tiny\textcolor{gray}{ $\uparrow$ 0.27}} & 95.22~{\tiny\textcolor{gray}{ $\uparrow$ 0.49}} & 92.31~{\tiny\textcolor{gray}{ $\downarrow$ 0.2}} & - \\
    \textbf{SMVCNN+Ours (S)} & 95.22~{\tiny\textcolor{gray}{ $\uparrow$ 1.37}} & 93.10~{\tiny\textcolor{gray}{ $\uparrow$ 1.74}} & 77.57~{\tiny\textcolor{gray}{ $\uparrow$ 0.92}} & 74.63~{\tiny\textcolor{gray}{ $\downarrow$ 0.2}} & 84.25~{\tiny\textcolor{gray}{ $\uparrow$ 0.52}} & 80.07~{\tiny\textcolor{gray}{ $\uparrow$ 0.25}} & 95.10~{\tiny\textcolor{gray}{ $\uparrow$ 0.37}} & 92.19~{\tiny\textcolor{gray}{ $\downarrow$ 0.32}} & \textbf{1h 43min} \\
    \arrayrulecolor{gray!50}\midrule 
    \arrayrulecolor{black}
    DAN \cite{2021dan} & 94.81 & 91.34 & 75.36 & 72.78 & 83.63 & 80.52 & 95.02 & 92.47 & 3h 24min \\
    \textbf{DAN+Ours (A)} & 95.08~{\tiny\textcolor{gray}{ $\uparrow$ 0.27}} & 93.05~{\tiny\textcolor{gray}{ $\uparrow$ 1.71}} & 76.35~{\tiny\textcolor{gray}{ $\uparrow$ 0.99}} & 71.07~{\tiny\textcolor{gray}{ $\downarrow$ 1.71}} & 84.09~{\tiny\textcolor{gray}{ $\uparrow$ 0.46}} & \textbf{81.31}~{\tiny\textcolor{gray}{ $\uparrow$ 0.79}} & \textbf{95.54}~{\tiny\textcolor{gray}{ $\uparrow$ 0.52}} & 93.19~{\tiny\textcolor{gray}{ $\uparrow$ 0.72}} & -\\
    \textbf{DAN+Ours (S)} & 94.67~{\tiny\textcolor{gray}{ $\downarrow$ 0.14}} & 93.51~{\tiny\textcolor{gray}{ $\uparrow$ 2.17}} & 76.96~{\tiny\textcolor{gray}{ $\uparrow$ 1.6}} & 73.67~{\tiny\textcolor{gray}{ $\uparrow$ 0.89}} & 83.63~{\tiny\textcolor{gray}{ $\uparrow$ 0}} & 80.21~{\tiny\textcolor{gray}{ $\downarrow$ 0.31}} & 94.98~{\tiny\textcolor{gray}{ $\downarrow$ 0.04}} & 93.04~{\tiny\textcolor{gray}{ $\uparrow$ 0.57}} & \textbf{2h 30min} \\
    \arrayrulecolor{gray!50}\midrule 
    \arrayrulecolor{black}
    VSFormer \cite{2024vsformer} & 93.44 & 89.49 & 75.97 & 73.98 & 83.63 & 80.11 & 95.26 & 92.89 & 2h 28min \\
    \textbf{VSFormer+Ours (A)} & 94.81~{\tiny\textcolor{gray}{ $\uparrow$ 1.37}} & 92.85~{\tiny\textcolor{gray}{ $\uparrow$ 3.36}} & 77.11~{\tiny\textcolor{gray}{ $\uparrow$ 1.14}} & 76.15~{\tiny\textcolor{gray}{ $\uparrow$ 2.17}} & 84.31~{\tiny\textcolor{gray}{ $\uparrow$ 0.68}} & 81.09~{\tiny\textcolor{gray}{ $\uparrow$ 0.98}} & 95.30~{\tiny\textcolor{gray}{ $\uparrow$ 0.04}} & 92.89~{\tiny\textcolor{gray}{ $\uparrow$ 0}} & - \\
    \textbf{VSFormer+Ours (S)} & 94.81~{\tiny\textcolor{gray}{ $\uparrow$ 1.37}} & 92.46~{\tiny\textcolor{gray}{ $\uparrow$ 2.97}} & 77.41~{\tiny\textcolor{gray}{ $\uparrow$ 1.44}} & 75.60~{\tiny\textcolor{gray}{ $\uparrow$ 1.62}} & \textbf{84.72}~{\tiny\textcolor{gray}{ $\uparrow$ 1.09}} & 81.05~{\tiny\textcolor{gray}{ $\uparrow$ 0.94}} & 95.50~{\tiny\textcolor{gray}{ $\uparrow$ 0.24}} & \textbf{93.27}~{\tiny\textcolor{gray}{ $\uparrow$ 0.38}} & \textbf{1h 47min} \\
    \bottomrule
    \end{tabular}
    }
    \label{tb:performance FG3D}
\end{table*}

\subsection{Interpretability} \label{sec:interpret}

We analyze the interpretability of Proto-FG3D from both global and local perspectives. The \textbf{interpretable class prototypes} are used to capture global 3D fine-grained shape characteristics, while the \textbf{view-level class activation maps (CAMs)} reveal local detail features.

\begin{table}[!htbp]
    \centering
    \tiny
    \renewcommand{\arraystretch}{0.8} 
    \caption{Samples and corresponding prototypes respectively assigned to classes by our Proto-FG3D model during the training and testing stages.}
    \resizebox{0.8\textwidth}{!}{ 
    \begin{tabular}{ccc|cccc}
    \toprule
    \multicolumn{3}{c|}{\textbf{Training}} & \multicolumn{4}{c}{\textbf{Testing}} \\
    \multicolumn{2}{c}{\textbf{Subcategory}} & \textbf{Prototype $\rightarrow$ Samples} &  \textbf{Test Sample} & \multicolumn{3}{c}{\textbf{Top-3 Prototypes}}\\
    \midrule
    \multirow[c]{2}{*}[-3mm]{\makecell{Majority class\\\textcolor{gray}{\ (train/test)}}} & \makecell{airliner\\\textcolor{gray}{\tiny (955/100)}} & \makecell{
    \includegraphics[width=0.03\textwidth]{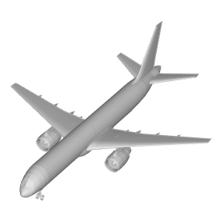}
    \hspace{0.5mm}
    \includegraphics[width=0.03\textwidth]{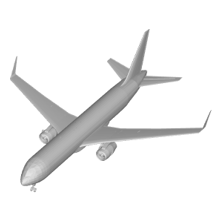}
    \hspace{0.5mm}
    \includegraphics[width=0.03\textwidth]{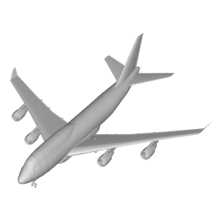}
    } & \makecell{\includegraphics[width=0.03\textwidth]{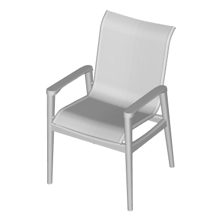}} & \makecell{\includegraphics[width=0.03\textwidth]{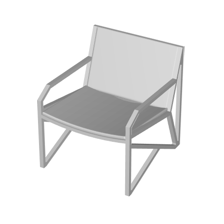}} & \makecell{\includegraphics[width=0.03\textwidth]{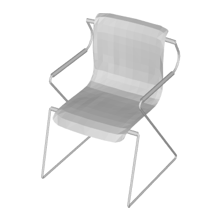}} & \makecell{\includegraphics[width=0.03\textwidth]{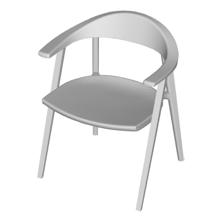}} \\
    & \makecell{helicopter\\\textcolor{gray}{\tiny (549/100)}} & \makecell{
    \includegraphics[width=0.03\textwidth]{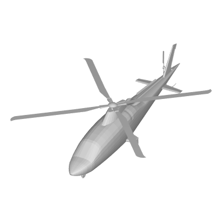}
    \hspace{0.5mm}
    \includegraphics[width=0.03\textwidth]{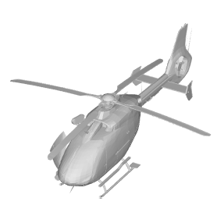}
    \hspace{0.5mm}
    \includegraphics[width=0.03\textwidth]{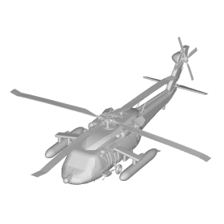}
    } & \makecell{Armchair\\Similarity} & \makecell{Armchair\\0.8779} & \makecell{Armchair\\0.4751} & \makecell{Armchair\\0.1714} \\
    \arrayrulecolor{gray!50}\midrule 
    \arrayrulecolor{black}
    \multirow[c]{2}{*}[-3mm]{\makecell{Minority class\\\textcolor{gray}{\tiny (train/test)}}} & \makecell{seaplane\\\textcolor{gray}{\tiny (35/20)}} & \makecell{
    \includegraphics[width=0.03\textwidth]{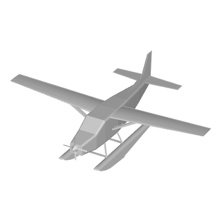}
    \hspace{0.5mm}
    \includegraphics[width=0.03\textwidth]{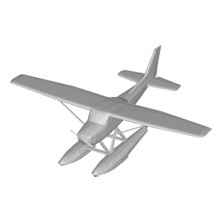}
    \hspace{0.5mm}
    \includegraphics[width=0.03\textwidth]{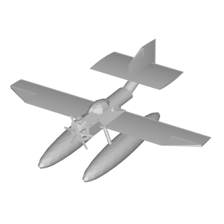}
    } & \makecell{\includegraphics[width=0.03\textwidth]{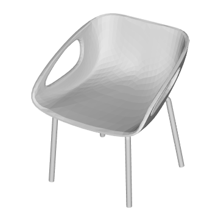}} & \makecell{\includegraphics[width=0.03\textwidth]{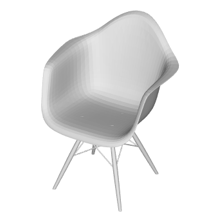}} & \makecell{\includegraphics[width=0.03\textwidth]{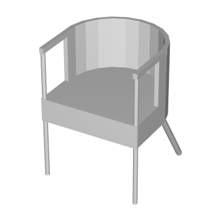}} & \makecell{\includegraphics[width=0.03\textwidth]{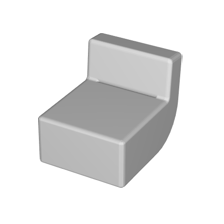}} \\
    & \makecell{transport\\\textcolor{gray}{\tiny (15/7)}} & \makecell{
    \includegraphics[width=0.03\textwidth]{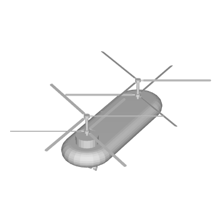}
    \hspace{0.5mm}
    \includegraphics[width=0.03\textwidth]{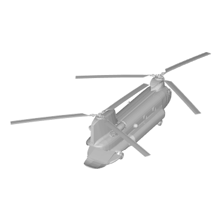}
    \hspace{0.5mm}
    \includegraphics[width=0.03\textwidth]{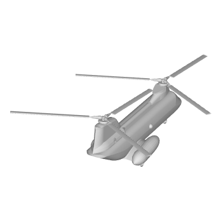}
    } & \makecell{Ball\\Similarity} & \makecell{Ball\\0.7941} & \makecell{Ball\\0.4590} & \makecell{Straight\\0.1431} \\
    \bottomrule
    \end{tabular}
    }
    \label{tb:anchor prototypes}
\end{table}

\noindent \textbf{Interpretable Class Prototypes.$\,$} Table \ref{tb:anchor prototypes} showcases representative prototypes selected from \textit{Airplane}, \textit{Car}, and \textit{Chair}. For each dataset, including the top-3 prototypes from both majority and minority subclasses for each dataset. 

The selected representative prototypes demonstrate significant diversity in both visual appearance and viewpoint variations, while maintaining clear discriminative patterns across subclasses in each dataset. Notably, the framework consistently extracts semantically meaningful prototypes even for minority subclasses, demonstrating its robustness in addressing class imbalance challenges common in fine-grained 3D recognition tasks. We also measure the similarity scores between input test samples and their assigned prototypes during inference. The top-matched prototypes consistently achieve high similarity values across both majority and minority classes, confirming robust alignment between the learned prototype representations and the feature distributions of test instances.

\noindent \textbf{View-Level CAMs.$\,$} Table \ref{tb:Views_cam} demonstrates the explainability of FG3D predictions, revealing local interpretability through CAM visualization. For each input sample, we present: (i) the top-5 representative views with activation maps, selected by view-prototype similarity scoring, and (ii) their corresponding prototypical parts. We observe that for majority classes such as \textit{Airliner} and \textit{Helicopter}, the selected views consistently capture key structural features (e.g., wing assemblies and rotor systems), highlighting the model's precise localization of discriminative class-specific regions. 

\begin{table*}[htbp]
\renewcommand{\arraystretch}{0.8}
\centering
\tiny
\caption{Sample explanation of predictions using top-5 representative views selected by Proto-FG3D with Grad-CAM.}
\resizebox{0.8\textwidth}{!}{
\begin{tabular}{c|ccccc|c}
\toprule
\textbf{Subcategory} & \multicolumn{5}{c}{\textbf{Top-5 Representative View-level Prototypes}} & \textbf{Prototypical Part} \\ 
\midrule
Airliner    & \makecell{
\includegraphics[width=0.05\textwidth]{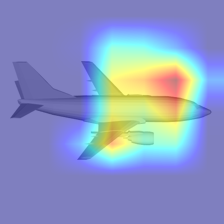}} &
\makecell{\includegraphics[width=0.05\textwidth]{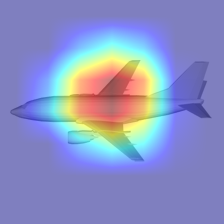}} &
\makecell{\includegraphics[width=0.05\textwidth]{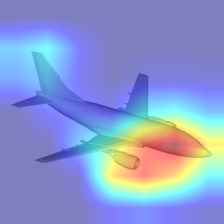}} &
\makecell{\includegraphics[width=0.05\textwidth]{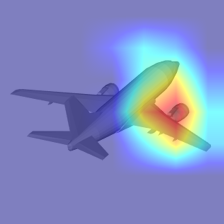}} &
\makecell{\includegraphics[width=0.05\textwidth]{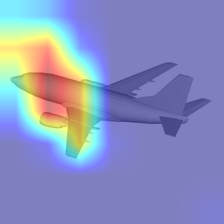}
} & \multirow[c]{2}{*}[2mm]{\makecell{
\includegraphics[width=0.05\textwidth]{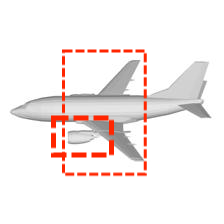}
}} \\
Similarity  & 0.6492 & 0.6054 & 0.5878 & 0.4653 & 0.3458 & \\
\hline
Helicopter  & \makecell{
\includegraphics[width=0.05\textwidth]{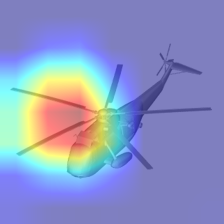}} &
\makecell{\includegraphics[width=0.05\textwidth]{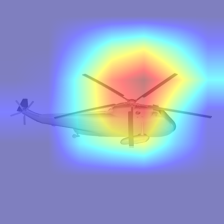}} &
\makecell{ \includegraphics[width=0.05\textwidth]{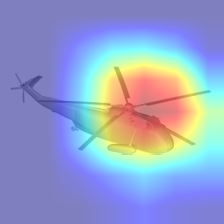}} &
\makecell{\includegraphics[width=0.05\textwidth]{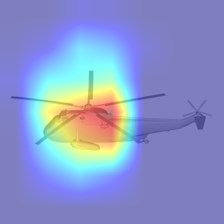}} &
\makecell{ \includegraphics[width=0.05\textwidth]{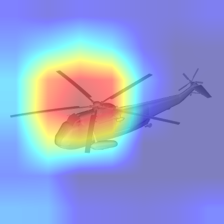}
} & \multirow[c]{2}{*}[2mm]{\makecell{
\includegraphics[width=0.05\textwidth]{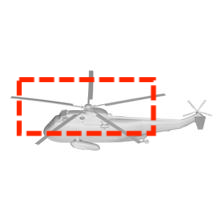}
}} \\
Similarity & 0.7413 & 0.5759 & 0.5685 & 0.5245 & 0.4846 \\ 

\bottomrule

\end{tabular}
}
\label{tb:Views_cam}
\end{table*}

\subsection{Ablation Study} 

\begin{table}[!htbp]
    \centering
    \caption{The ablation study of (a) loss function, (b) prototype number, (c) momentum coefficient, and (d) loss weight on Proto-FG3D performance for FG3D-Airplane dataset.}
    \label{tb:ablation_studies}
    \tiny 
    \begin{minipage}{0.21\textwidth}
        \centering
        \setlength{\tabcolsep}{1.5mm} 
        (a) Loss function. \\[0.2em]
        \begin{tabular}{cc|cc}
        \toprule
        $\mathcal{L}_{ce}$ & $\mathcal{L}_{pc}$  & AIA (\%) & ACA (\%) \\
        \midrule
        \ding{51} & \ding{55} & 94.67 & 92.38 \\
        \ding{51} & \ding{51} & \textbf{95.36} & \textbf{92.95} \\
        \bottomrule
        \end{tabular}
    \end{minipage}\hfill
    \begin{minipage}{0.21\textwidth}
        \centering
        \setlength{\tabcolsep}{1.5mm}
        (b) Prototype number. \\[0.2em]
        \begin{tabular}{c|cc}
        \toprule
        $K$ & AIA (\%) & ACA (\%) \\
        \midrule
        10 & 94.81 & 93.15 \\
        15 & 94.67 & 92.72 \\
        \textbf{20} & \textbf{95.36} & \textbf{92.87} \\
        25 & 94.95 & 92.77 \\
        30 & 95.08 & 92.56 \\
        \bottomrule
        \end{tabular}
    \end{minipage}\hfill
    \begin{minipage}{0.21\textwidth}
        \centering
        \setlength{\tabcolsep}{1.5mm}
        (c) Momentum coefficient. \\[0.2em]
        \begin{tabular}{c|cc}
        \toprule
        $\eta^0$ & AIA (\%) & ACA (\%) \\
        \midrule
        0 & 95.08 & 92.51 \\
        0.9 & 95.08 & 92.77 \\
        0.99 & 95.08 & 93.23 \\
        \textbf{0.999} & \textbf{95.63} & \textbf{93.95} \\
        0.9999 & 94.95 & 92.62 \\
        \bottomrule
        \end{tabular}
    \end{minipage}\hfill
    \begin{minipage}{0.21\textwidth}
        \centering
        \setlength{\tabcolsep}{1.5mm}
        (d) Loss weight. \\[0.2em]
        \begin{tabular}{c|cc}
        \toprule
        $\alpha$ & AIA (\%) & ACA (\%) \\
        \midrule
        0.05 & 94.26 & 92.15 \\
        0.10 & 95.36 & 92.87 \\
        0.15 & 94.40 & 91.67 \\
        \textbf{0.20} & \textbf{95.63} & \textbf{93.95} \\
        0.25 & 95.22 & 93.28 \\
        \bottomrule
        \end{tabular}
    \end{minipage}
\end{table}

We conduct comprehensive ablation studies on the FG3D-Airplane dataset using GVCNN as the backbone network. As summarized in Table~\ref{tb:ablation_studies}, the prototype contrastive loss $\mathcal{L}_{pc}$ boosts 3D fine-grained classification by enhancing feature discriminability (95.36\% AIA). Optimal prototype count ($K$=20) balances intra-class compactness and inter-class separation. A high momentum coefficient ($\eta^0$=0.999) ensures stable prototype evolution and feature consistency. Balanced loss weighting ($\alpha$=0.20) harmonizes cross-entropy guidance and contrastive discrimination, achieving peak performance while preventing training instability. 

\section{Conclusion}
In this paper, we proposed Proto-FG3D, the first prototype-based framework for fine-grained 3D shape classification. Departing from conventional softmax-based classifiers, Proto-FG3D employs prototype learning to simultaneously: (i) enhance inter-subcategory discrimination through online clustering that dynamically updates view-specific prototype representations, (ii) mitigate class imbalance via adaptive prototype allocation, and (iii) provide built-in interpretability through visualizable prototypes. Experimental results on both FG3D (fine-grained) and ModelNet40 (standard) benchmarks demonstrate that Proto-FG3D achieves state-of-the-art classification accuracy while providing comprehensive interpretability through global prototype visualization and local view-activation mapping. In future work, we will investigate adaptive prototype evolution, leverage large-scale FG3D datasets and explore unsupervised or self-supervised techniques to broaden the applicability of Proto-FG3D.

\bibliographystyle{unsrt}  
\bibliography{mybibfile}

\end{document}